# Seeing Through the Fog

## A Cost-Effectiveness Analysis of Hallucination Detection Systems

Alexander Thomas  
Wisecube AI

Seth Rosen  
Wisecube AI

Vishnu Vettrivel  
Wisecube AI

Abstract:

This paper presents a comparative analysis of hallucination detection systems for AI, focusing on automatic summarization and question answering tasks for Large Language Models (LLMs). We evaluate different hallucination detection systems using the diagnostic odds ratio (DOR) and cost-effectiveness metrics. Our results indicate that although advanced models can perform better they come at a much higher cost. We also demonstrate how an ideal hallucination detection system needs to maintain performance across different model sizes. Our findings highlight the importance of choosing a detection system aligned with specific application needs and resource constraints. Future research will explore hybrid systems and automated identification of underperforming components to enhance AI reliability and efficiency in detecting and mitigating hallucinations.

## 1. Introduction

We will explore the realm of Large Language Model (LLM) hallucinations, exploring their nature and characteristics. We will shed light on the systems employed to detect these hallucinations, delving into their inner workings and examining how they can be compared to each other. Furthermore, we will investigate the multifaceted landscape of metrics available for evaluating the efficacy of such detection systems.

Next, we will embark on a journey through the evaluation methodology, scrutinizing the various datasets utilized to conduct a comprehensive comparison among hallucination detection systems. This exploration will provide invaluable insights into the practical implementation and assessment of these systems.

Subsequently, we will unravel the results yielded by the diverse hallucination detection systems, uncovering their strengths and weaknesses. This analysis will enable us to discern the most effective approaches for identifying and mitigating LLM hallucinations.

To conclude, we will engage in a discussion of the trade-offs associated with different detection systems. This will include an examination of their respective advantages and disadvantages, guiding future research and development efforts. Additionally, we will propose potential avenues for further improvements, aiming to refine and enhance the capabilities of hallucination detection systems. Through this comprehensive exploration, we aspire to deepen our understanding of LLM hallucinations and contribute to the advancement of detection systems capable of safeguarding the integrity and reliability of language models.

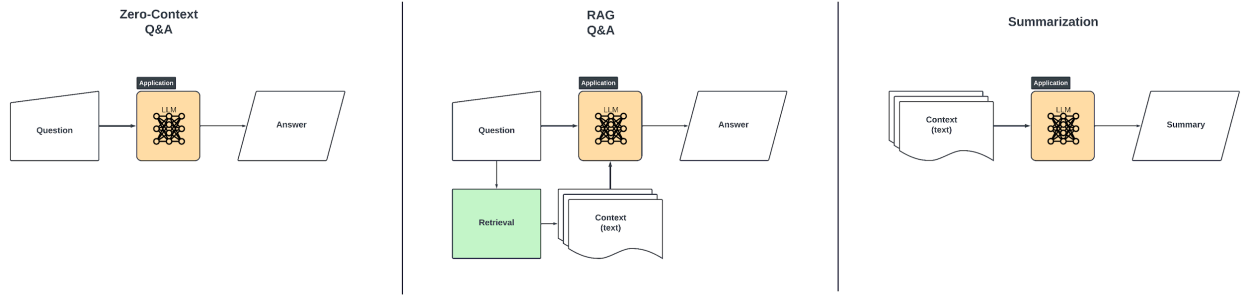

| Zero-Context Q&A | RAG + Q&A | Summarization |
|---|---|---|
| Small inputs, Variable-sized outputs<br>No external information | **R**etrieval **A**ugmented **G**eneration<br>Small inputs + contexts = Large Inputs<br>Variable-sized outputs | Larger inputs<br>Outputs smaller than inputs<br>More complex concept of accuracy |
| `{"question": "Is fexofenadine efficacious and safe in children ( aged 6-11 years ) with seasonal allergic rhinitis?",`<br><br>`"answer": "Yes. The efficacy and safety of the H(1)-antihistamine fexofenadine has been confirmed in this multicenter, multinational study of children aged 6 to 11 years with seasonal allergic rhinitis."}` | `{"question": "Rennae Stubbs and Renáta Tomanová were both what?",`<br>`"context": "Rennae Stubbs (born 26 March 1971) is an Australian retired tennis player.Renáta Tomanová (born 9 December 1954) is a former professional tennis player from Czechoslovakia.",`<br><br>`"answer": "tennis player"}` | `{"summary": "Plans to mark the 50th anniversary of the oldest carnival in europe have been announced.",`<br><br>`"reference": "Plans have been announced for the "oldest carnival in europe" to celebrate its 50th anniversary. The west indian carnival in leeds was launched in 1967 and once again will take over potternewton park..."}` |

Figure 1.The different kinds systems categorized by the context provided in the task.

Large Language Models (LLMs) represent an important breakthrough in natural language processing (NLP) technology. Firstly, their ability to generate large amounts of fluent text [Brown, T. et al., 2020], then their application as a natural language interface for a wide set of use cases like programming [Zhang, Y. et al., 2024] and image generation [Minaee, S. et al., 2024], among others. Just as the abilities are novel, so are the problems. In this paper we will be talking about the hallucination problem. Hallucinations are incorrect, but often believable, statements made by an LLM [Banerjee, S. et al., 2024]. If an application exposes direct LLM output to a user, this creates a number of dangers. Depending on the domain of the application, this could be a danger to the user, liability to the application owner, or embarrassment to the application owner. We will be comparing two hallucination detection systems. The first system is Pythia, and the second is LynxQA [Ravi, S. S. et al., 2024], secondarily we will also compare them with a simplistic system as well called Grading. The differences between these systems let us explore the different design choices when building or adopting such a system. We will focus on the automatic summarization and retrieval augmented generation for question answering (RAG-QA) use case, that is, we will be measuring the hallucination detection system on datasets from these two use cases. We will show how the diagnostic odds ratio metric can be adopted for this task, as well as looking at quality-cost tradeoffs.

## 2. Background and Related Work

In Large Language Models (LLMs), hallucinations are "nonsensical or unfaithful" text generated by the model, deviating from reality or context. They fall into two categories: [Ji, Z. et al., 2023]

1. **Intrinsic Hallucinations**: These directly contradict the provided source material. Examples include generating factually incorrect statements or illogical conclusions.
2. **Extrinsic Hallucinations**: These hallucinations don't directly contradict the source but introduce unverifiable elements. This could involve speculation or claims that cannot be confirmed against the source.

### 2.1. The Cost of Hallucinations

Hallucinations in AI technology can cause significant financial losses and safety issues. For instance, Google lost $100 billion due to a hallucination in a promotional video.[Coulter, M., & Bensinger, G., 2023]. Companies can be liable for chatbot claims. Air Canada's chatbot offered a discount, but the airline denied it. The court ruled in favor of the customer. [Moffatt v. Air Canada, SC-2023-005609, 2024]

> *27. Air Canada argues it cannot be held liable for information provided by one of its agents, servants, or representatives – including a chatbot. It does not explain why it believes that is the case. In effect, Air Canada suggests the chatbot is a separate legal entity that is responsible for its own actions. This is a remarkable submission. While a chatbot has an interactive component, it is still just a part of Air Canada's website. It should be obvious to Air Canada that it is responsible for all the information on its website. It makes no difference whether the information comes from a static page or a chatbot.*
> *28. I find Air Canada did not take reasonable care to ensure its chatbot was accurate. While Air Canada argues Mr. Moffatt could find the correct information on another part of its website, it does not explain why the webpage titled "Bereavement travel" was inherently more trustworthy than its chatbot. It also does not explain why customers should have to double-check information found in one part of its website on another part of its website.*

### 2.2. Related Work

The most reliable form of hallucination detection is still human labeling [Ji, Z. et al., 2023]. However this is not tenable for measuring at scale, let alone monitoring. Classic metrics like ROUGE, BLEU, and METEOR are ineffective for detecting errors in generated text as they penalize different language styles.t [Ji, Z. et al., 2023, Zha, Y. et al., 2023]. There are also some approaches that attempt to use non-LLM models (e.g. BERT) to detect hallucinations [Zhang, T. et al., 2020, Zha, Y. et al., 2023]. LLMs are more generalizable than these models for this task. Some systems directly "ask" the model if the given example is a hallucination, this is known as the LLM-as-a-judge [Ravi, S. S. et al., 2024]. Although this seems very simplistic, when combined with other techniques, like fine-tuning, it can be effective [Ravi, S. S. et al., 2024]. Some other approaches attempt to verify information given in the LLM output [Es, S. et al., 2023]. The difficulties deciding between these systems is compounded by the variety of metrics used in evaluating them.

## 2.3. Metrics

Currently, there is no single metric that is considered standard. The novelty of this issue means that there has not been enough time to coalesce. Different systems and authors adopt different metrics depending on what data they have available. Some use accuracy [Ravi, S. S. et al., 2024], some use Spearman correlation [Zha, Y. et al., 2023], etc. We considered three approaches to metrics in our work. First, we can use the metric already associated with the dataset being used. This has the benefit of us not needing to re-interpret the labels into some other form, for example, binarizing them, as well as allowing us to compare our hallucination detection system to other kinds of systems evaluated on the dataset. The downsides are that there is no ability to generalize the results into a score to compare different systems, and some of the dataset-associated metrics are not amenable to the hallucination detection task. The second approach is to binarize the labels and use accuracy, [Ravi, S. S. et al., 2024]. This has the benefit of being easy to calculate, and it requires effectively no explanation for someone to understand the metric. The downside is that accuracy is not a robust metric, it is very sensitive to imbalanced data. The third option is to binarize the labels but choose a more sophisticated metric, such as diagnostic odds ratio. This has the benefit of being independent of prevalence. The downside is that the metric does require some explanation. Also, this metric is not bound by 0 and 1, instead it is bound on the lower end by 0, and is unbound on the higher end.

The diagnostic odds ratio (DOR) measures the effectiveness of a diagnostic test by combining sensitivity and specificity into a single value. A DOR of 1 indicates no discrimination, while higher values indicate better diagnostic accuracy. It's used in medical research to compare and evaluate diagnostic tests.

[Glas, A. S. et al., 2003]. In our case, we are "diagnosing" the hallucinatory text.

A. $$\text{Diagnostic odds ratio, DOR} = \frac{TP/FN}{FP/TN} = \frac{TP/FP}{FN/TN} = \frac{TP \cdot TN}{FP \cdot FN}$$

B. $$\text{SE}(\ln \text{DOR}) = \sqrt{\frac{1}{TP} + \frac{1}{FN} + \frac{1}{FP} + \frac{1}{TN}}$$

C. $$\ln \text{DOR} \pm 1.96 \times \text{SE}(\ln \text{DOR})$$

Figure 2. DOR formulae, A score definition where TP, FP, TN, FN stand for true positives, false positives, true negatives, false negatives respectively; B standard error of log DOR; C 95% confidence interval for log DOR [Afina S Glas et al., 2003]

Cost is crucial for hallucination detection systems as the task can be complex. Latency may not be directly significant, but it can define the cost if the application owner hosts the model. For interventions, latency becomes separately important.

## 3. Methodology

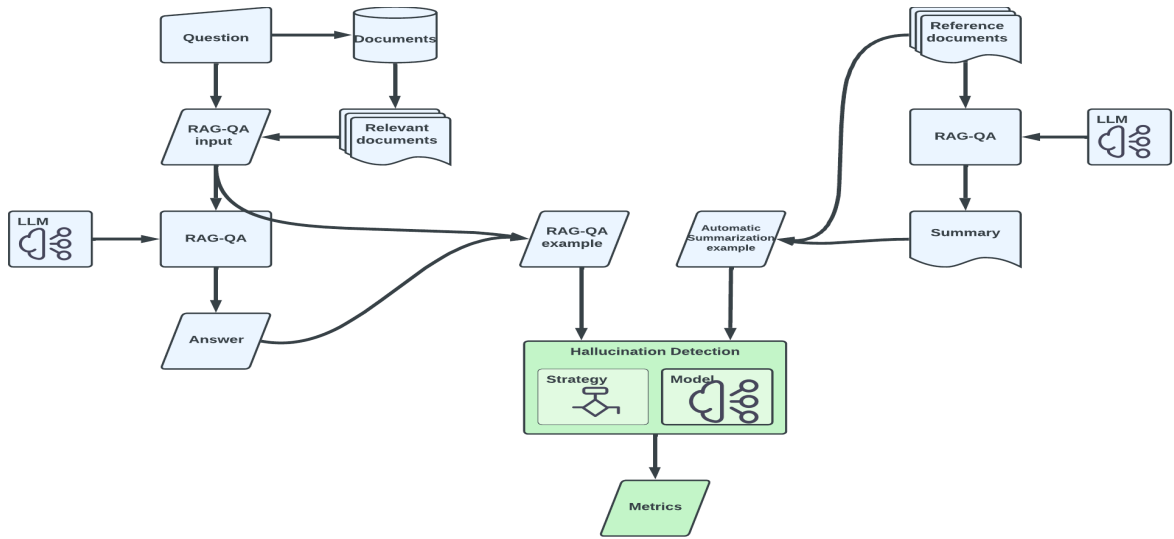

Figure 3. Diagram of RAG-QA and Automatic Summarization systems and how they plug into a hallucination detection system

### 3.1 Data

The two tasks selected for this experiment were selected due to them being common LLM application tasks, and there being many viable datasets. The automatic summarization task consists of 2174 examples from three different data sources. Two of these data sources use the same CNN / Daily Mail data set as references, but the summaries were not generated in the same manner.

The data sets making up the automatic summarization data set are: QAGS-CNNDM, QAGS-XSUM and SummEval. These data sets were selected from the work done on AlignScore [Zha, Y. et al., 2023]. The QAGS-CNNDM data set contains news articles from CNN and Daily Mail. The summaries are generated from a QA summarization process. The labels are from crowdsourced human labelers [Wang, A. et al., 2020]. The QAGS-XSUM is made from news articles from the BBC, but otherwise is similar to QAGS-CNNDM [Wang, A. et al., 2020]. The SummEval data set is also made from news articles from CNN and Daily Mail, but the summaries are generated from various models. The labels for SummEval come from both experts and crowdsourced human labels with these labels marked separately in the source data [Fabbri, A. R. et al., 2021]. Since there is substantial disagreement between the expert labels and the crowdsourced labels, the expert labels are used. These data sets have the advantage of being classic NLP data sets. This means that the data is well understood by the NLP community. The data sets making up the RAG-QA data set are: DROP, FinanceBench, RAGTruth, and TruthfulQA. The first three of these data sets were selected since they are part of HaluBench used by LynxQA [Ravi, S. S. et al., 2024].

The other three data sets from HaluBench were excluded. PubmedQA and covidQA were excluded due to some data quality issues experienced by the team in this and previous projects. HaluEval was excluded since its hallucinatory answers were produced by asking an LLM to hallucinate [Li, J. et al., 2023]. In the team's opinion, this produces low quality examples. The fourth data set in the RAG-QA data set is TruthfulQA. The TruthfulQA data set tests language models' truthfulness with 817 questions across 38 categories, focusing on common misconceptions. It evaluates how well models avoid generating misleading or incorrect answers, emphasizing factual accuracy. This data set is structured such that each row has a question, an ideal answer, good answers, bad answers, and a source reference. Most of the source references are Wikipedia and other URLs. Where these URLs could be resolved they were used to create the context. This left 688 rows. The data set was then exploded along the answers yielding 6735 examples. The data sets from HaluBench were selected since we were testing LynxQA. TruthfulQA was added since it allowed us to test LynxQA on new data.

Each system is evaluated against data sets, and DOR and cost metrics are calculated. We compare these metrics to baselines for performance and cost.

**3.2 Overview of Strategies Tested in Hallucination Detection**

In the pursuit of improving the reliability of AI systems, particularly in natural language generation tasks, various strategies for hallucination detection have been developed and tested. This section introduces the key strategies evaluated in our comparative analysis, highlighting their mechanisms and intended applications. Each strategy leverages distinct approaches to mitigate the impact of hallucinations, enabling AI developers to select the most suitable method based on their specific use cases.

When we discuss systems in this paper we are referring to a combination of a *strategy* and a *model*. The strategy can be a singular prompt, a chain, or a more complex program for evaluation. The model is generally a LLM model, but could also be an ensemble of LLMs, or even a completely different kind of model. In comparing Pythia and LynxQA we are primarily comparing their strategies. We will also look at their performance on other models as well.

**1. LynxQA Strategy**

LynxQA uses large language models (LLMs) to generate and verify answers to questions. While effective, it relies on the model's accuracy and requires continuous fine-tuning, leading to potentially high costs.

## 2. Pythia Strategy

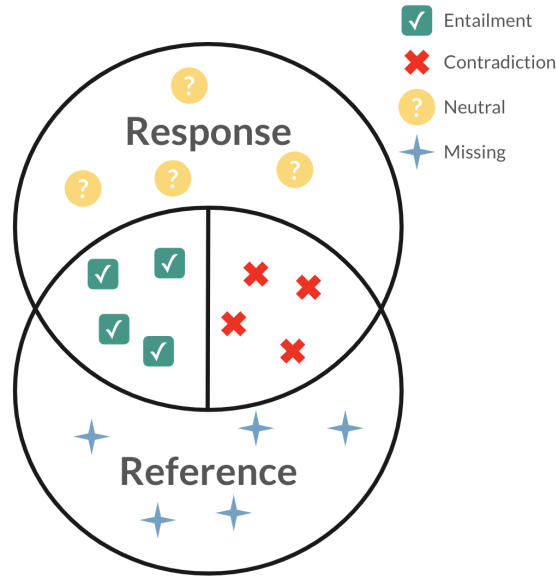

Figure 4.. Venn diagram of the categorization of claims used by Pythia. Entailment is for claims supported in the reference, contradiction for claims refuted in the reference, neutral for claims in the response but not the reference, and missing for claims in the reference but not the response.

Pythia is a proprietary strategy based on the concept of claim extraction. The idea is that correctness or faithfulness of an application's output is approximately equivalent to how well supported the constituent claims of the output are supported by the context.
We separate use cases into different kinds of context and different kinds of outputs. Automatic summarization has large context (references) and large outputs, though the outputs should be smaller than the provided references. RAG-QA has large context and variable outputs. Whenever a piece of input or output is large, it is viable for claim extraction. Otherwise, it will be treated as a claim itself. Pythia uses the classic natural language inference classes of *entailment, contradiction, neutral* to classify the claims in the output [Silva, V.. et al., 2018]. From this, a factual accuracy metric is calculated. This factual accuracy score is then thresholded to make a prediction (diagnosis).

$$\text{Accuracy} = \frac{w_e + w_c + w_r}{w_e(\text{Entailment} + \epsilon)^{-1} + w_c(1 - \text{Contradiction} + \epsilon)^{-1} + w_r(\text{Reliability} + \epsilon)^{-1}}$$

Figure 5. Factual accuracy metric extraction-based techniques, where *Entailment* is the proportion of claims that are supported by reference or context, *Contradiction* is the rate of claims refuted, and *Reliability* is an optional parameter reflecting the quality of claims categorized as Neutral.

**Algorithm 1** Pythia Summarization Evaluation Strategy

**Input**
$S$ generated summary
$R$ concatenated reference text
$Q$ optional question

**Parameters**
$M$ model
$E$ extractor
$Ch$ checker

**Output**
$T_s$ triples extracted from $S$
$C$ summary triple classifications
$A$ factual accuracy score

**function** EVALUATE_SUMMARY($S$: text, $R$: text, $Q$: Option[text])
    $T_s \leftarrow E.\text{extractor}(S, M)$     ▷ summary triples
    $T_r \leftarrow E.\text{extractor}(R, M)$     ▷ reference triples
    **for** $t_{S,i} \in C_s$ **do**
        $c_i \leftarrow Ch.\text{check}(t_{S,i}, T_r)$
    **end for**
    $E \leftarrow$ proportion of summary claims that are supported by reference
    $C \leftarrow$ proportion of summary claims that are refuted by reference
    $A \leftarrow \frac{2}{(E+\epsilon)^{-1} + (1-C-\epsilon)^{-1}}$
    **return** $C, T_s, A$
**end function**

Figure 6. This pseudocode defines the Pythia strategy. Given a generated summary, concatenation of the reference texts, and an optional question, extract the summary and reference triples, then classify the summary triples, finally computing the factual accuracy of the summary

Pythia distributes task complexity between LLM, extractor, and checker. This approach provides more robustness and modularity against hallucinations and maintains performance across model sizes. This flexibility makes Pythia a promising choice for hallucination detection.

### 3. Baseline Strategy - Grading

The Grading strategy is a simple one-prompt approach intended to work with minimal fine tuning. It instructs the LLM to grade the response using the A-F US grading system. This works due to the many examples of text qualitatively judged with such a grade. Although the commonness of this kind of evaluation helps the LLM assign a grade, the examples the model might have seen likely are not purely about factual correctness. So there is potential for a non-hallucinatory response to be given a low grade to some quality problem other than accuracy (e.g. writing style). The opposite problem can also occur if a hallucinatory response has otherwise good qualities.

Each strategy operates under unique principles that define their approach to detecting hallucinations:

**LynxQA** relies on the strength of the LLM alone, emphasizing prompt engineering, but at the cost of increased resource consumption for model improvements.

**Pythia** enhances detection capabilities by integrating LLM outputs with external logic, allowing for a distributed approach to processing and error mitigation.

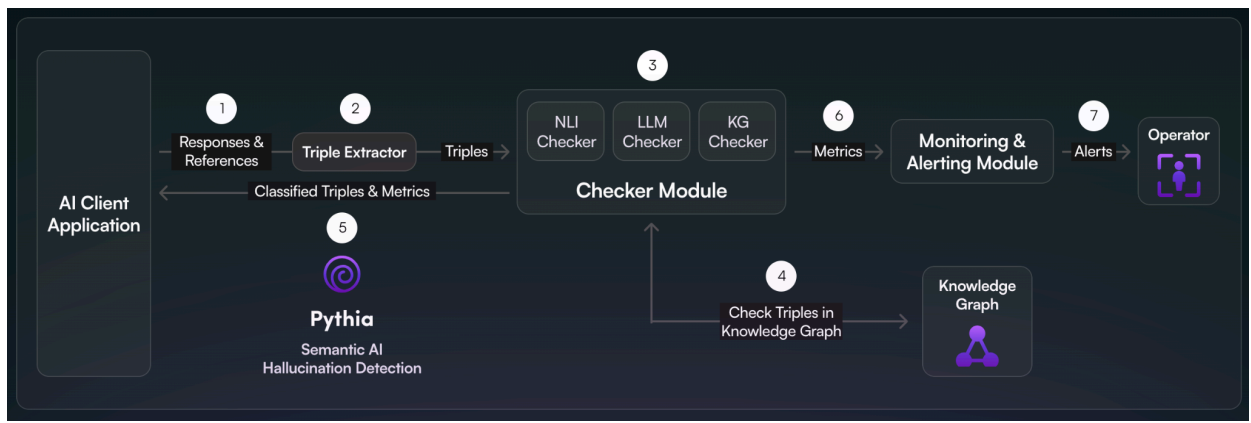

Figure 7. Diagram of Pythia deployed and integrated with an AI client application.

**Baseline Strategy** capitalize on the diversity of models to foster more reliable outputs through validation, yet they require sophisticated orchestration.

By comparing these strategies, our study aims to identify their strengths and weaknesses in real-world applications, providing valuable insights for developers looking to enhance the reliability of their AI systems in the face of inherent challenges like hallucinations.

## 3.1. Case Study

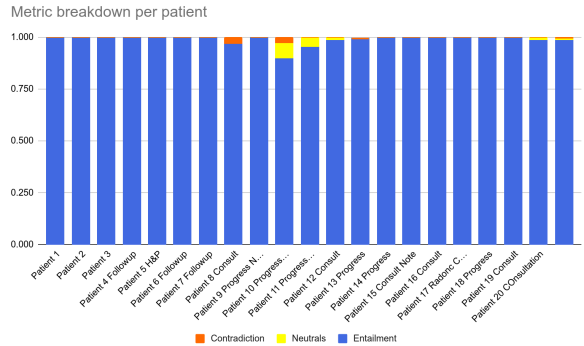

Figure 8. This bar chart shows the distribution of claim classifications by client's test documents. Blue is entailment, yellow is neutral, and red-orange is contradiction.

The primary goal of this Pythia case study was to enhance the trustworthiness of generative AI outputs for a major pharmaceutical partner. The focus was on reducing the occurrence of hallucinations to ensure precision in patient care and safety. The pharmaceutical industry faces significant challenges related to the accuracy of AI-generated content. With some LLMs exhibiting error rates exceeding 25% in their outputs, the implications of these inaccuracies can be severe, including operational disruptions, reputational damage, and life-threatening risks in patient care. Their existing methods of addressing these hallucinations relied on human annotation, which proved insufficient for scalability and timely intervention.

Pythia's automated evaluation of patient summaries generated by large language models (LLMs) showed a strong correlation with human-annotated results, validating the system's effectiveness in identifying inaccuracies. However, despite these positive findings, the occasional detection of discrepancies underscores the need for continuous monitoring to maintain reliability over time.

## 4. Results and Analysis

Firstly, we will look over the automatic summarization results. To understand the performance, first let's look at the distribution of PASS/FAIL by component data set. All three source data sets are highly imbalanced. The labels were binarized by looking for perfect grades, and complete label agreement.

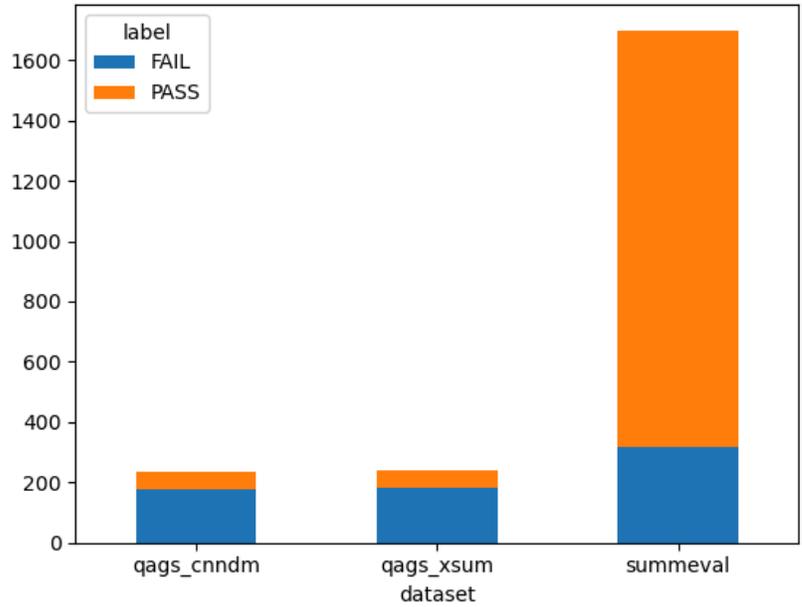

Figure 9. distribution of PASS/FAIL by component data set in the automatic summarization data set

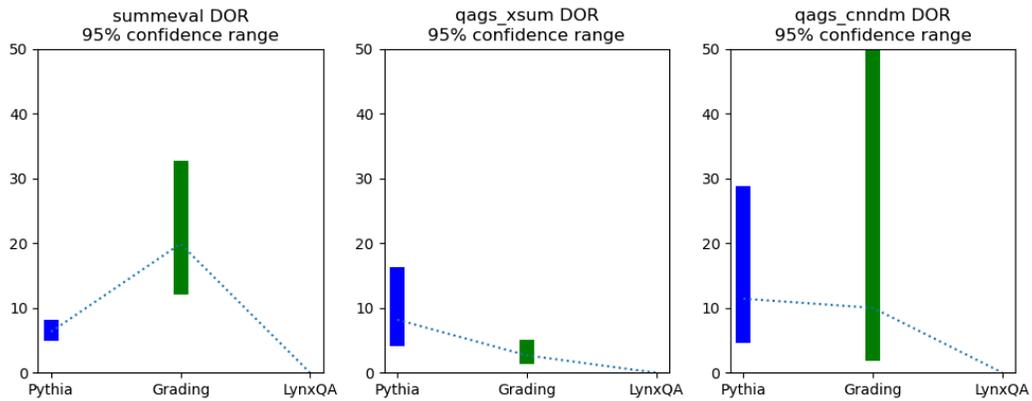

Figure 10. DOR score (95% confidence interval) across the automatic summarization component datasets

Since LynxQA is only defined for the RAG-QA task, it does not have a score here. We see that The Grading strategy performs best on SummEval, but Pythia does better on the QAGs data. Note also that the size of the 95% confidence interval for Grading is quite large on the QAG CNNDM data set. The interpretation of the score 11.44 (Pythia QAGS-CNNDM) is the ratio of the odds that a summary is assigned "FAIL" given that it is labeled "FAIL, with respect to the odds that a summary is assigned "FAIL" given that it is labeled "PASS".

Now let us look at the RAG-QA data. With the exception of RAGTruth these data sets are nearly balanced. The difference between these data sets and the automatic summarization datasets illustrates why it is important to have a metric that is independent of prevalence.

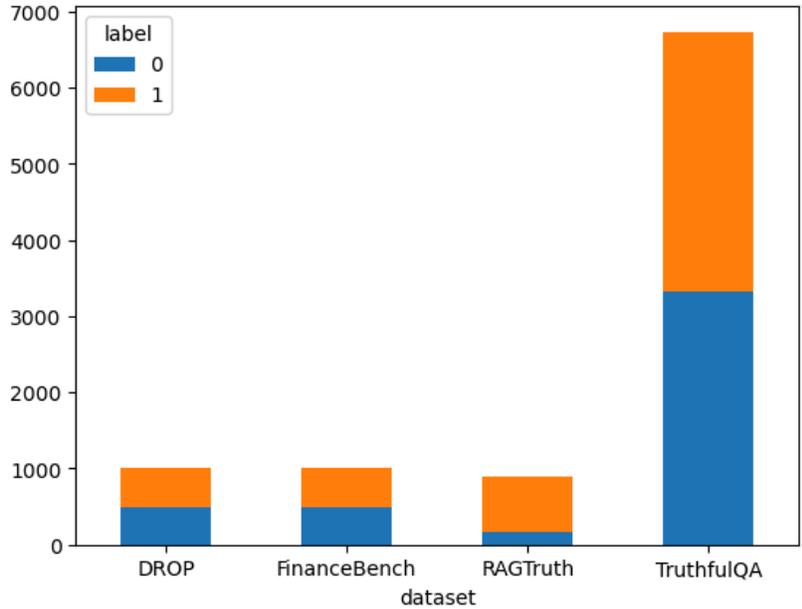

Figure 11. distribution of PASS/FAIL by component data set in the RAG-QA data set

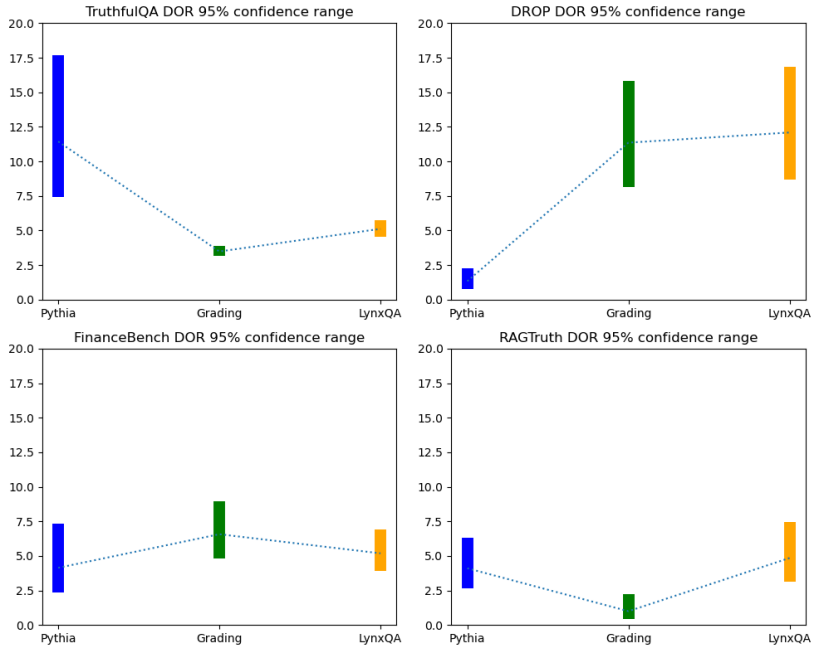

Figure 12. DOR score (95% confidence interval) across the RAG-QA component datasets

With the RAG-QA data, LynxQA performs competitively with the datasets on which it was engineered. Pythia performs best on the TruthfulQA dataset. The Grading strategy performs similarly to LynxQA on all datasets except RAG-QA.

The three strategies compared above were all run with the GPT-4o-mini model. The LynxQA strategy was also run with the GPT-4o model, since it performs similarly to the fine tuned model [Ravi, S. S. et al., 2024]. This run performed well in terms of its DOR metric, but it incurs 16.85 times the cost.

Finally, let's compare these strategies using the Grading strategy as a baseline.

| Strategy | Model | DOR over Grading | Cost over Grading ($1/1M tokens) |
| --- | --- | --- | --- |
| lynxqa | gpt-4o | 1.91 | 16.85 |
| lynxqa | gpt-4o-mini | 0.71 | 1.00 |
| grading | gpt-4o-mini | 1.00 | 1.00 |
| pythia | gpt-4o-mini | 1.28 | 1.00 |

Table 1. Comparing the ratio systems performance and cost to Grading

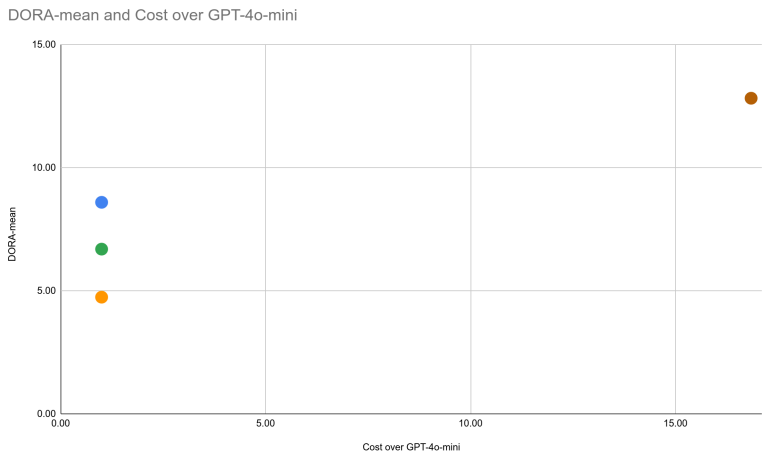

Figure 13. Scatter plot of Table 1.

## 4.1 Analysis

In the analysis of the results, we can see that the difference between Pythia and LynxQA is less, relatively speaking, on the TruthfulQA dataset. This dataset is the fairest comparison between the two, since it is the task that LynxQA supports, but it was not prompt engineered on. Here we see that the difference between Pythia and LynxQA (GPT-4o), for DOR over Grading, is actually slightly worse, 50% higher ratio for overall data vs 30% higher for TruthfulQA in isolation.

| Strategy | Model | DOR over Grading | Cost over GPT-4o-mini |
|---|---|---:|---:|
| lynxqa | gpt-4o | 4.30 | 16.85 |
| lynxqa | gpt-4o-mini | 1.47 | 1 |
| grading | gpt-4o-mini | 1 | 1 |
| pythia | gpt-4o-mini | 3.28 | 1 |

Table 2. Same comparison to Table 1, but only on TruthfulQA

In general, someone who is building an LLM application will need to know the parameters of their product. In considering which measurement approach is best, the developer must consider what resources they have to invest. If cost is not an obstacle, something like LynxQA can be a baseline to building a highly customized model that is as accurate as possible. One still would need to consider what this cost would mean for the final product. Alternatively, Pythia offers the benefit of being able to use less expensive LLMs with the tradeoff being a moderate increase in complexity.

Users can interpret these results by examining the specific metrics reported, such as the DOR metric and the performance benchmarks against other methods. This allows them to assess not only the effectiveness of the hallucination detection system but also to identify areas for improvement within their AI models. The ability to pinpoint specific issues—such as consistent errors in certain contexts or tasks—enables developers to refine their AI systems proactively. Consequently, this insight empowers organizations to make data-driven decisions that enhance the reliability and performance of their AI applications, ultimately fostering greater user confidence and compliance with industry standards.

## 5.Conclusion & Future Direction

**Conclusion**

The findings of this paper illustrate that the LynxQA strategy demonstrates improved performance when paired with a more advanced model, resulting in a higher diagnostic odds ratio (DOR). However, this improvement comes at a significantly increased cost, reflecting the inherent design differences between strategies. LynxQA utilizes a single prompt that relies entirely on the language model (LLM) to address the problem's complexity. In contrast, Pythia integrates the LLM with external logic, demonstrating only a minor performance decrease when transitioning from GPT-4o to GPT-4o-mini in previous research.

In any LLM-based application, developers face a critical decision regarding the extent to which they depend on the LLM to address the problem. This includes determining whether to use the LLM for the entire solution or to employ it for specific components while leveraging other models, algorithms, or heuristics for different parts. The trade-off lies in resource allocation versus potential improvements. Strategies that heavily rely on LLMs can achieve straightforward enhancements through fine-tuning specialized models. However, this reliance also poses risks;

if the task complexity demands a larger model, costs can escalate dramatically. Conversely, employing an LLM for subtasks results in a more robust system across various model families and sizes, although improvements may become more complicated as it requires pinpointing and addressing underperforming components within the system.

When choosing an effective hallucination detection system, it is essential to consider the specific applications being evaluated. For instance, if an application encompasses various tasks like automatic summarization and retrieval-augmented question answering (RAG-QA), the detection system must be versatile enough to identify hallucinations across both functions. This consideration of versatility is crucial in making informed trade-offs between strategy and model selection.

**Future Research Directions**

To build upon the findings of this paper, future research could focus on developing hybrid systems that optimize the balance between LLM reliance and external logic integration, enhancing versatility and cost-effectiveness. Additionally, investigating methods to automate the identification of underperforming components within complex systems may lead to more efficient strategies for improvement. Further studies could also explore the long-term impacts of different detection mechanisms on diverse applications, providing a comprehensive understanding of how to best deploy and refine hallucination detection systems in real-world scenarios.

# Appendix

## Definition of Terms

| Term | Definition |
| --- | --- |
| **Automatic Summarization** | The process of creating a concise and coherent summary of a larger text while retaining the essential information. |
| **Bias** | Systematic favoritism or discrimination present in AI model predictions |
| **Cost Metrics** | Measures used to evaluate the financial implications of deploying a detection system, including the costs and operational efficiency. |
| **Diagnostic Odds Ratio (DOR)** | A statistical measure indicating the odds of detecting a true positive relative to the odds of false positives in detection systems. |
| **Evaluation Framework** | A structured approach for assessing the performance, accuracy, and reliability of AI models and their outputs based on metrics & standards. |
| **Explainability** | The degree to which the internal decision-making processes of an AI system can be understood and interpreted by humans. |
| **Hallucination** | In the context of language models, it refers to the generation of plausible but factually incorrect or misleading content by the model. |
| **Latency Metrics** | Measurements that assess the delay or response time of a system |
| **LLM (Large Language Model)** | A type of artificial intelligence model trained on a vast corpus of text data to understand and generate human-like text. |
| **Model Drift** | The gradual degradation of a model's performance over time. |
| **Pythia** | AI observability platform designed to detect hallucinations & improve the reliability of AI systems through continuous monitoring and analysis. |
| **Question Answering (QA)** | A natural language task where the system retrieves data from a knowledge base relevant to questions posed in natural language to use in answering the question. |
| **Subtasks** | Individual components or specific tasks that are part of a larger process or problem-solving strategy, often delegated to various models or algorithms within a system. |